\documentclass{article}

     \PassOptionsToPackage{numbers, compress}{natbib}

\usepackage[main, final]{neurips_2025}

\usepackage{framed}
\usepackage{graphicx}
\usepackage[utf8]{inputenc} %
\usepackage[T1]{fontenc}    %
\usepackage[hidelinks]{hyperref}       %
\usepackage{url}            %
\usepackage{booktabs}       %
\usepackage{amsfonts}       %
\usepackage{nicefrac}       %
\usepackage{microtype}      %
\usepackage{xcolor}         %

\workshoptitle{Efficient Reasoning}

\title{Agentic NL2SQL to Reduce Computational Costs}

\author{
  Dominik Jehle \\
  \texttt{jehled@cs.uni-freiburg.de}\\
  University of Freiburg \\
  Freiburg im Breisgau, Germany
  \And
  Lennart Purucker \\
  \texttt{purucker@cs.uni-freiburg.de}\\
  University of Freiburg \\
  Freiburg im Breisgau, Germany
  \And
  Frank Hutter\\
  Prior Labs\\
  ELLIS Institute Tübingen\\
  University of Freiburg\\
}

\begin{document}

\maketitle

\begin{abstract}
Translating natural language queries into SQL queries (NL2SQL or Text-to-SQL) has recently been empowered by large language models (LLMs). Using LLMs to perform NL2SQL methods on a large collection of SQL databases necessitates processing large quantities of meta-information about the databases, which in turn results in lengthy prompts with many tokens and high processing costs. To address this challenge, we introduce Datalake Agent, an agentic system designed to enable an LLM to solve NL2SQL tasks more efficiently. Instead of utilizing direct solvers for NL2SQL that call the LLM once with all meta-information in the prompt, the Datalake Agent employs an interactive loop to reduce the utilized meta-information. Within the loop, the LLM is used in a reasoning framework that selectively requests only the necessary information to solve a table question answering task. We evaluate the Datalake Agent on a collection of 23 databases with 100 table question answering tasks. The Datalake Agent reduces the tokens used by the LLM by up to 87\% and thus allows for substantial cost reductions while maintaining competitive performance.
\end{abstract}

\section{Introduction}
Large language models (LLMs) show significant potential to advance the field of information retrieval and understanding of tabular data. Current models have demonstrated notable improvements in tasks related to the understanding of tabular data \citep{zhu_large_2024, liu_survey_2024, shi_survey_2024}.
Optimizing the prompts plays a critical role in maximizing the performance of models and can significantly influence the output behavior for a specific task \citep{memon_llm-informed_2024, cheung_reality_2024}.
At the same time, the number of tokens to represent the prompt directly determines the cost of a request \citep{openai_pricing_nodate, deepseek_models_nodate}. Reducing the number of tokens decreases the overall cost, which is often needed as large prompts can quickly entail significant expenses.

In the field of translating natural language queries into SQL queries (NL2SQL or Text-to-SQL) \citep{liu_survey_2024}, prompt size can quickly explode when the number of SQL databases grows. 
This issue becomes particularly relevant when working in large enterprise companies that store an extensive amount of structured, tabular data.
Typically, when solving tasks on large collections of databases, LLMs receive large amounts of meta-information about the structure and types of all databases in one direct prompt. Yet, most NL2SQL tasks require only a subset of the database, rendering much of the meta-information in the prompt useless while still incurring additional costs for users.
\\
Thus, we propose to select only the meta-information truly necessary for a single task to significantly reduce the cost of solving the NL2SQL task. 
To this end, we introduce Datalake Agent, an agentic system designed to enable an LLM to solve NL2SQL tasks more efficiently by reducing the utilized meta-information through a reasoning framework. The framework operates across three core areas: information acquisition, iterative refinement, and query formulation.

We evaluate the Datalake Agent by comparing it to a direct prompting strategy with OpenAI’s GPT-4-mini \citep{openai2024gpt4omini}.
Both methods, the Datalake Agent and the direct prompting, are evaluated on a set of 319 tables, collected from 23 different databases, some of which contain real-world data \citep{robinson_relbench_2024}. Both methods are tested on a manually created collection of 100 table question answering tasks.
Our results show that the Datalake Agent significantly reduces the costs of processing large amounts of data. Furthermore, the Datalake Agent retains competitive performance on average, while also improving the LLM's capability to process complex queries more effectively.

\textbf{Our contributions are:}  
(1) the \textbf{Datalake Agent}, a reasoning framework that enables large language models (LLMs) to efficiently navigate and query large volumes of meta-information;  
(2) a \textbf{new benchmark of 100 table-based question answering tasks}, carefully curated to evaluate model performance across varying levels of complexity.

\section{Related Work}

There are various methods for leveraging large language models (LLMs) for large data. One possible method is to provide the LLM direct access to all data by transmitting all information via one prompt to the model.
However, \citet{qiu_tqa-bench_2024} have shown a clear negative performance trend across all tested models: when the prompt size increases, the overall performance declines. Likewise, \citet{dong_large_2024} observed that for prompts exceeding $1000$ tokens, the effectiveness of GPT models for tabular data devolves to mere guesswork.
Thus, alternatives have been developed, such as Text2API \citep{wang_redefining_2024} or NL2SQL \citep{liu_survey_2024}.
In Text2API, instead of granting the model direct access to complete table information, this approach enables interaction with the dataset through API endpoints \citep{wang_redefining_2024}.
While this method demonstrates impressive results, it restricts the model to retrieving information exclusively through the available API endpoints and cannot access data beyond them.
Instead, translating natural language queries into SQL (NL2SQL) offers two major advantages. First, there is no need to input entire tables into the model. Instead, the model receives only meta-information about the tables, which is especially beneficial for datasets with a large number of rows. Second, unlike Text2API, NL2SQL does not impose limitations on information retrieval. By generating SQL queries, the model can access and process all the necessary data dynamically \citep{liu_survey_2024}.
While LLMs show remarkable capabilities in translating natural language queries into SQL, their accuracy decreases with growing schema complexity, larger numbers of tables, and more demanding query structures \citep{liu_survey_2024}.
Our work focuses on improving the efficiency of NL2SQL with LLMs when there is an excessive amount of meta-information available, while maintaining or enhancing their performance.
\\
The community developed several benchmarks for NL2SQL, such as the Bird \citep{li2023can, li2025swe} and Spider \citep{DBLP:journals/corr/abs-1809-08887,lei2024spider20evaluatinglanguage, lei2024spider} benchmarks. 
Yet these benchmarks only provide the relevant, potentially necessary, meta-information to models instead of all the available meta-information.
Our work establishes a new benchmark, building on prior benchmarks. We simulate a potentially more realistic scenario in an enterprise company where the user, and by extension the LLM, do not know which database contains the necessary meta-information.
Instead, the LLM is tasked with solving an NL2SQL task, specifically a table question answering task, given the natural language query and the meta-information of all available databases and tables.

\section{Method}
To efficiently handle large volumes of meta-information, we employ the \textbf{Datalake Agent}, a reasoning framework that guides large language models (LLMs) through structured, task-driven information acquisition. The method can be explained in terms of three core areas:\\
\textbf{Information Acquisition.}\quad 
In the information acquisition, the LLM begins by gathering general schema knowledge. The Datalake Agent provides an intermediary layer for structured exploration using predefined commands: \texttt{GetDBDescription} retrieves high-level summaries of available databases, \texttt{GetTables} enumerates tables within a selected database, and \texttt{GetColumns} exposes column-level metadata, including names and types. A visual overview of the workflow is provided in the Appendix \ref{The Datalake Agent}.\\
\textbf{Iterative Refinement.}\quad    
During iterative refinement, the LLM follows a hierarchical approach, progressing from broad, high-level data toward task-specific details. At any point, it can revert to a coarser level of information if needed, before refining again. This flexible, feedback-driven loop allows the LLM to reason autonomously over complex database structures while ensuring that only relevant information is retrieved.\\
\textbf{Query Formulation.}\quad    
Finally, in query formulation, once sufficient schema information has been collected, the LLM generates precise SQL queries using \texttt{DBQueryFinalSQL}. The Datalake Agent executes these queries through a dedicated access layer, which guarantees scalability and modularity. Integrating new databases requires minimal adjustments while preserving the systematic reasoning approach. This hierarchical and iterative method allows the LLM to autonomously select relevant information, refine its understanding, and generate accurate queries.\\

\section{Experimental Setup}

\textbf{Model.}\quad
For our experiments, we employ the OpenAI GPT-4 Mini model \citep{openai2024gpt4omini}, accessed via the OpenAI API. To ensure consistent behavior, the temperature parameter is fixed at $0.1$. The model outputs follow the structured response format described above.
\\  
\textbf{Databases.}\quad 
As the foundation for evaluation, we select five databases from the Relational Deep Learning Benchmark (RelBench) \citep{robinson_relbench_2024}. Since enterprise environments typically consist of large, heterogeneous database convolutes, we expand this collection with $18$ additional simulated databases from domains such as sports, politics, business, and geography. These simulated databases contain only schema information but no data, serving to increase the complexity of the evaluation setup.\\  
\textbf{Table Question Answering Benchmark.}\quad 
To evaluate the Datalake Agent, we construct a benchmark of $100$ natural language Table Question Answering tasks. Each task requires the model to generate a SQL query without being informed which tables are relevant. Each of the five RelBench databases contributes $20$ tasks, spanning two levels of difficulty.\\  
We evaluate the system under three experimental settings, each differing in the number of available databases and tables. The first setting includes only the RelBench databases, totaling $42$ tables. The second and third settings incorporate the simulated databases, resulting in $159$ and $319$ tables, respectively. Examples of benchmark questions are presented in Appendix \ref{Example Tasks for Evaluation}.\\
\textbf{Direct Prompt Baseline.}\quad 
For comparison, we design a direct prompt baseline that provides the model with all necessary schema information upfront. Its system prompt mirrors that of the Datalake Agent but excludes the option to request schema information incrementally.

\section{Results}

\begin{figure}[h]
    \centering
    \begin{minipage}{0.48\textwidth}
        \centering
        \includegraphics[width=\linewidth]{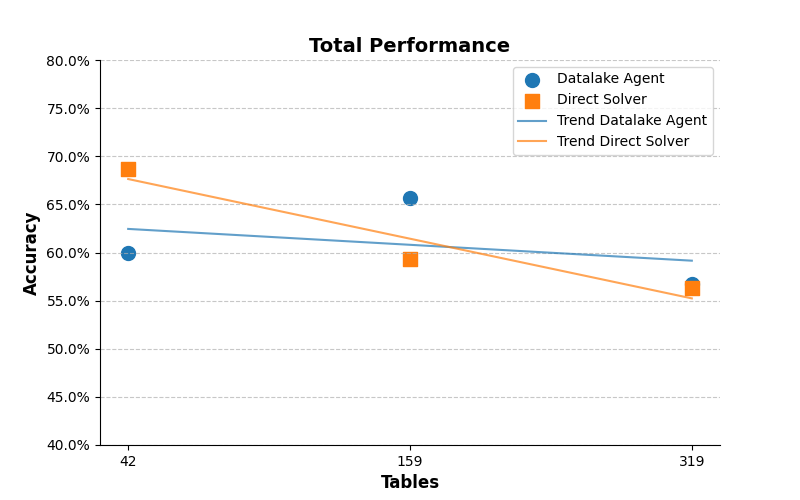}
        \caption{Performance trend of Direct Solver and Datalake Agent.}
        \label{fig:trend_performance}
    \end{minipage}
    \hfill
    \begin{minipage}{0.48\textwidth}
        \centering
        \includegraphics[width=\linewidth]{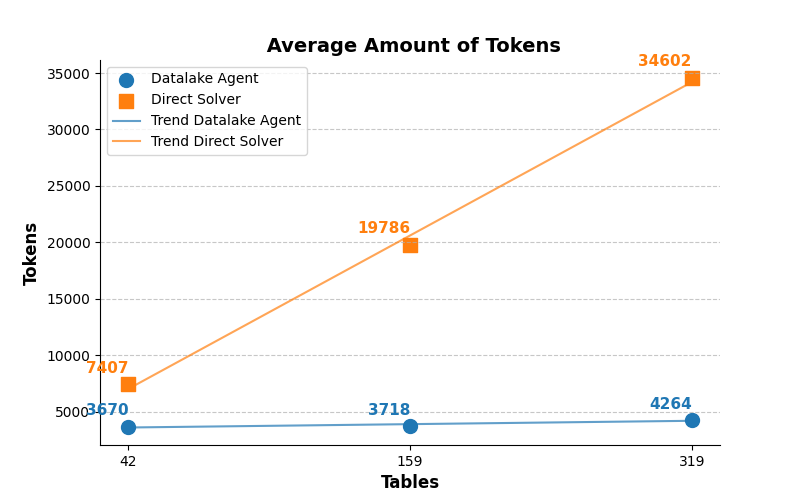}
        \caption{Trend of average amount of tokens required for solving one task.}
        \label{fig:trend_tokens}
    \end{minipage}
\end{figure}

\textbf{Performance.}\quad
Examining the trends of both approaches, as illustrated in Figure~\ref{fig:trend_performance}, reveals that the Direct Solver initially outperforms the Datalake Agent. However, as data volume increases, both methods experience a decline in accuracy. The Direct Solver exhibits a steeper drop in performance compared to the Datalake Agent, a trend that is particularly pronounced for more complex tasks. A more detailed performance analysis is provided in Appendix \ref{Performance}.

\textbf{Input Tokens.}\quad 
Figure~\ref{fig:trend_tokens} illustrates the average number of tokens consumed by the Datalake Agent and the Direct Solver as the number of tables increases. The Datalake Agent shows relatively stable token usage, rising slightly from 3,670 tokens with 42 tables to 4,264 tokens with 319 tables. In contrast, the Direct Solver exhibits a steep increase from 7,407 tokens at 42 tables to 34,602 tokens at 319 tables. These results indicate that the Datalake Agent scales more efficiently, maintaining consistent overhead even as database size grows, while the Direct Solver’s token consumption increases nearly linearly with the number of tables. Additional information on input token usage can be found in Appendix \ref{Input Tokens}.

\begin{figure*}[h]
    \centering
    \includegraphics[width=1\textwidth]{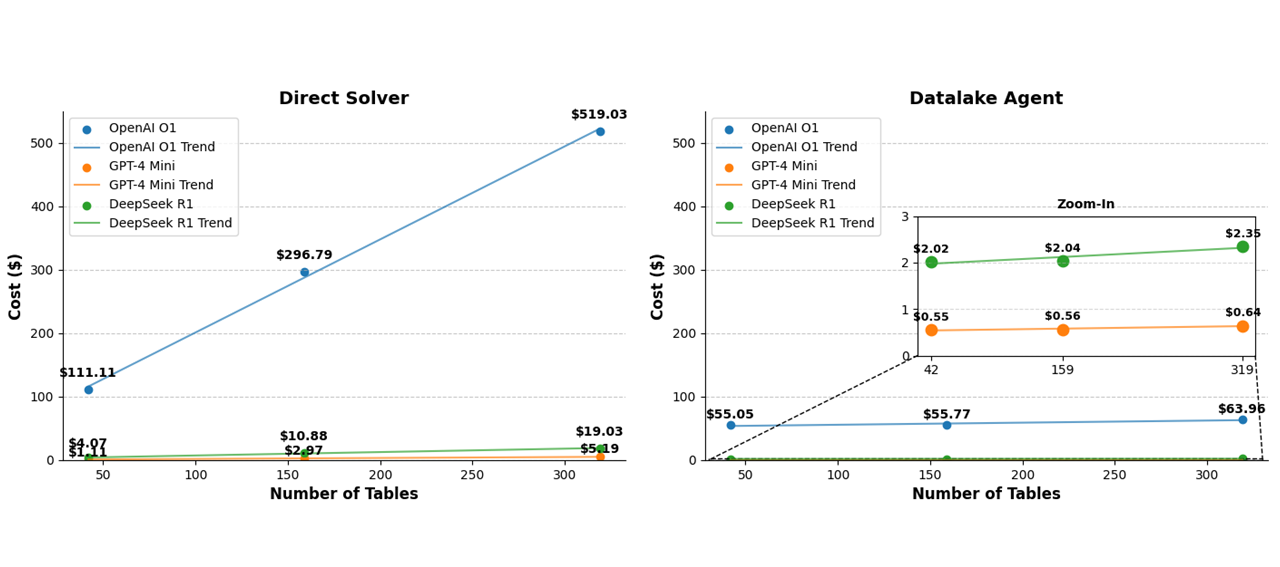}
    \caption{Comparison of costs for 1000 tasks across methods and LLMs.}
    \label{fig:costs}
\end{figure*}

\textbf{Associated Costs.}\quad
To contextualize token usage, Figure~\ref{fig:costs} presents a cost calculation based on average input tokens per task, scaled to 1000 tasks. API costs were taken from official company specifications \citep{openai_pricing_nodate,deepseek_models_nodate}. A significant difference emerges between the Direct Solver and the Datalake Agent. For a table size of 42, the Direct Solver incurs double the cost of the Datalake Agent, increasing to eight times the cost at 319 tables. For the OpenAI o1 model, this corresponds to a difference exceeding \$450 per 1000 tasks.

\section{Conclusion} 
This study investigates the impact of large-scale databases on LLM performance using NL2SQL approaches. A total of 23 databases with 319 tables, including both real and simulated data, were employed. To handle the substantial information volume, the Datalake Agent was introduced, enabling the model to selectively query only the data required for each task.

Evaluation on 100 Table-QA tasks, divided into simple (single-table) and complex (multi-table) tasks, across three settings with 42, 159, and 319 tables shows that while the Direct Solver performs well on smaller datasets, its performance deteriorates sharply as table numbers increase. In contrast, the Datalake Agent consistently outperforms the direct approach in larger and more complex settings, while requiring significantly fewer input tokens, thereby reducing computational costs.

\textbf{Limitations.} 
A key limitation of our agentic system is the potential for infinite reasoning loops. In some cases, the model struggles to identify the correct tables, leading to repeated requests and incorrect answers. A more detailed explanation is provided in the Appendix \ref{Handling of Infinite Reasoning Loops}.  Moreover, the evaluation is confined to GPT-4 Mini and only a few database settings, which may limit generalizability to broader scenarios.

\textbf{Future Work.} 
Future research should extend evaluations to larger and more complex datasets to assess scalability and generalization more thoroughly. Addressing the infinite loop issue is critical, as developing strategies to prevent repeated queries could improve reliability and accuracy. Further studies could explore additional LLMs and more diverse task types to fully understand the potential and limitations of the Datalake Agent in real-world applications.

\textbf{To conclude,} our Datalake Agent is a first step towards improving the efficiency of NL2SQL in enterprise use cases. Based on our promising first results, we believe the Datalake Agent can positively impact the use of LLMs for NL2SQL.

\section*{Acknowledgements}
L.P. acknowledges funding by the Deutsche Forschungsgemeinschaft (DFG, German Research Foundation) under SFB 1597 (SmallData), grant number 499552394;
Frank Hutter acknowledges the financial support of the Hector Foundation.
Finally, we thank the reviewers for their constructive feedback and contribution to improving the paper.
\bibliography{refs}

\newpage
\appendix
\section{The Datalake Agent}
\label{The Datalake Agent}
\begin{figure}[h] %
    \centering
    \includegraphics[width=1\textwidth]{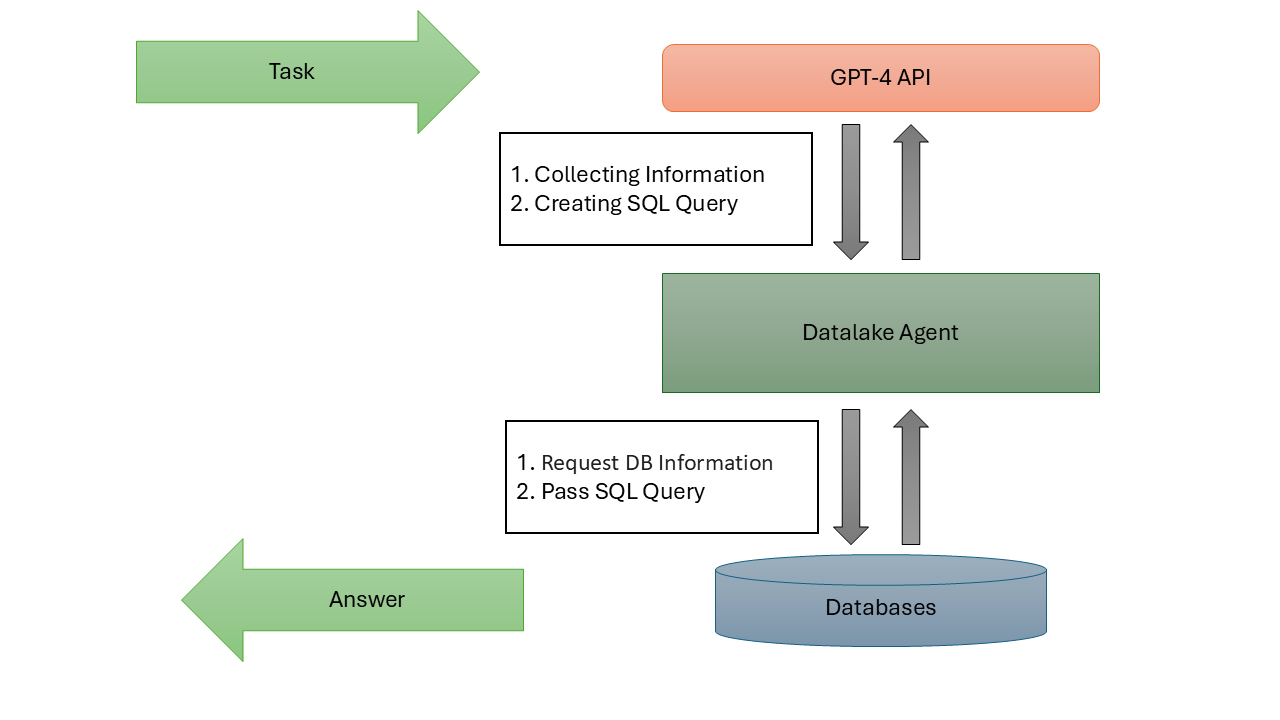} 
    \caption{Solution process of a Table Question Answering Task using the Datalake Agent}
    \label{fig:chatAgend} %
\end{figure}
\begin{figure}[h] %
    \centering
    \includegraphics[width=1\textwidth]{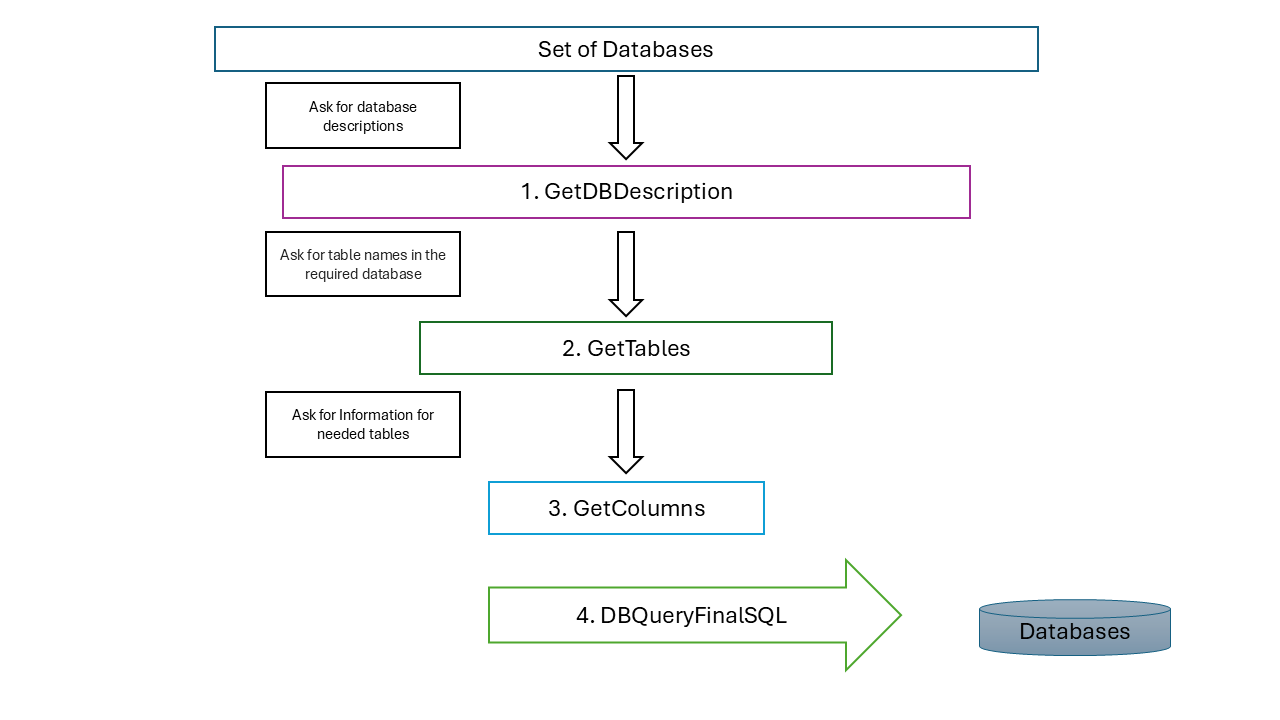} %
    \caption{Communication loop before solving a task between model and Datalake Agent}
    \label{fig:meinbild} %
\end{figure}
\newpage
\section{Input Tokens}
\label{Input Tokens}
\begin{figure}[!htbp] %
    \centering
    \includegraphics[width=0.8\textwidth]{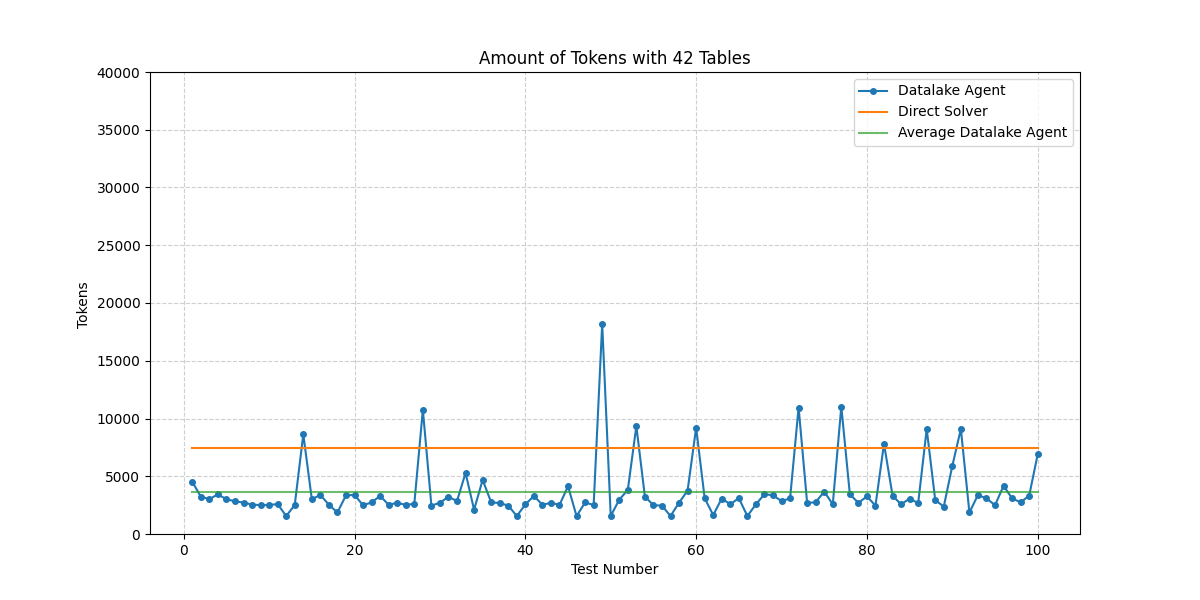}
    \caption{Required input tokens per test with 42 tables: Comparison of model using Datalake Agent and Direct Solver}
    \label{fig:tokens42}
\end{figure}

\begin{figure}[!htbp]
    \centering
    \includegraphics[width=0.8\textwidth]{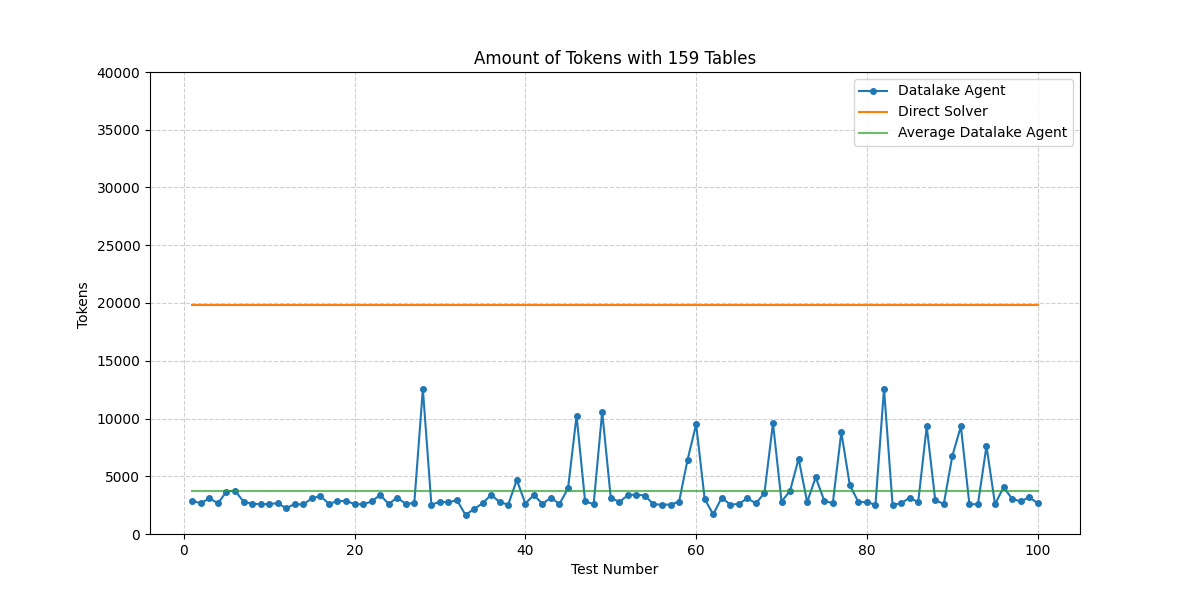}
    \caption{Required input tokens per test with 159 tables: Comparison of model using Datalake Agent and Direct Solver}
    \label{fig:tokens159}
\end{figure}

\begin{figure}[!htbp]
    \centering
    \includegraphics[width=0.8\textwidth]{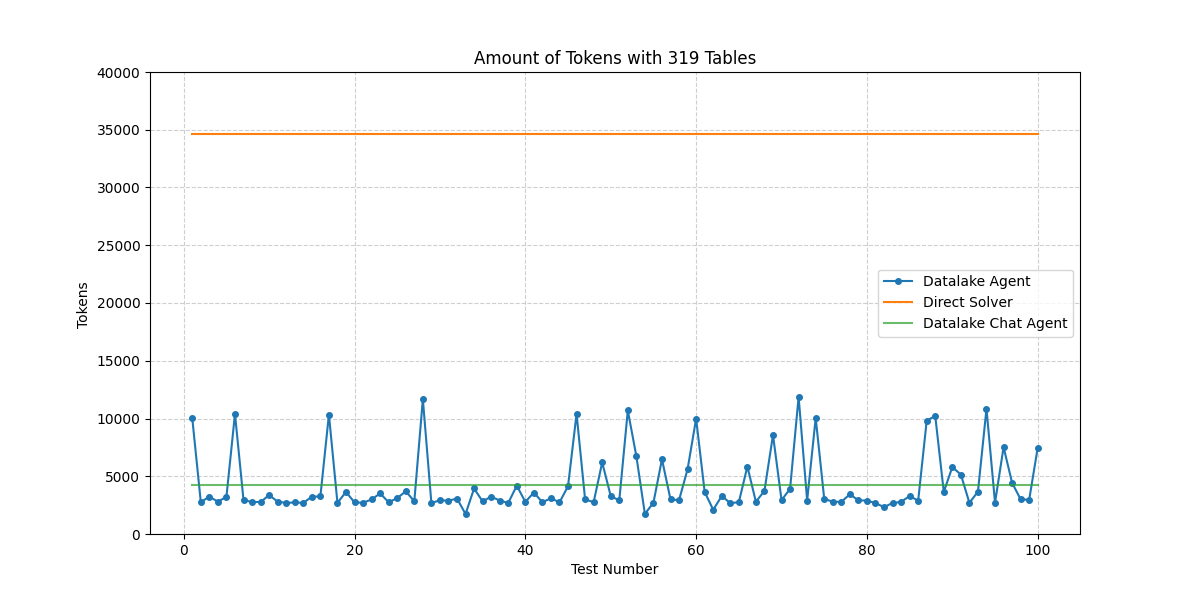}
    \caption{Required input tokens per test with 319 tables: Comparison of model using Datalake Agent and Direct Solver}
    \label{fig:tokens319}
\end{figure}
\section{Performance}
\label{Performance}
\begin{figure}[!htbp] %
    \centering
    \includegraphics[width=0.5\textwidth]{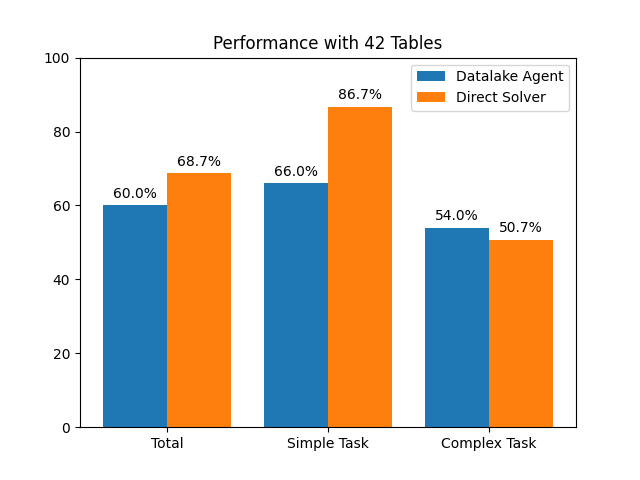}
    \caption{Correct answers to simple and more complex tasks with 42 tables}
    \label{fig:balken42}
\end{figure}

\begin{figure}[!htbp]
    \centering
    \includegraphics[width=0.5\textwidth]{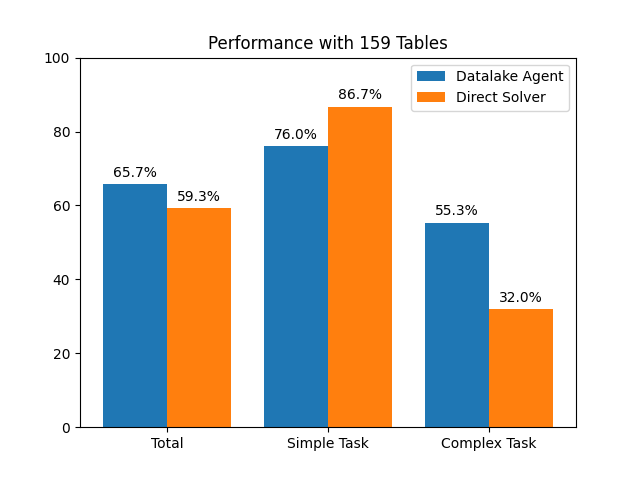}
    \caption{Correct answers to simple and more complex tasks with 159 tables}
    \label{fig:balken159}
\end{figure}

\begin{figure}[!htbp]
    \centering
    \includegraphics[width=0.5\textwidth]{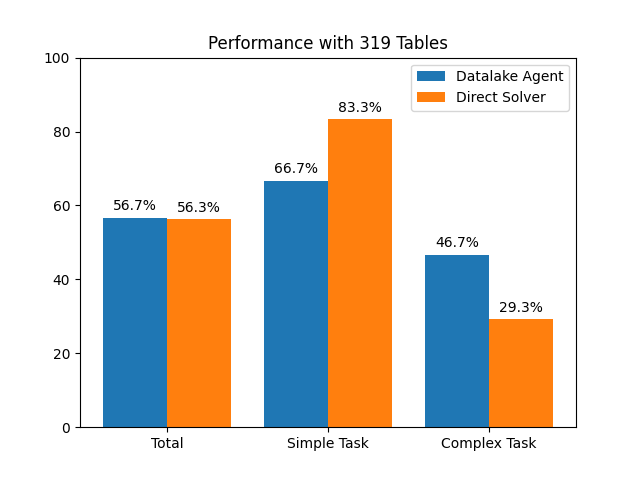}
    \caption{Correct answers to simple and more complex tasks with 319 tables}
    \label{fig:balken319}
\end{figure}
\newpage
\section{Example Tasks for Evaluation}
\label{Example Tasks for Evaluation}
\begin{framed}
1 Which driver has the most wins in Formual 1?\\
2 Give me the latest search query on Avito.\\
3 Give me the titles of the 2 most recent clinical trials.\\
4 Give me the total number of posts in Stack-Exchange.\\
5 How many clinical studies are available?\\
7 How many comments does the most frequently commented post on Stack Exchange have?\\
8 How many searches did Avito have on April 28. 2015?\\
9 Name 3 clinical trial sponsors.\\
10 How many Formula 1 races were there in 2015?\\
11 Give me the ids of the 10 newest posts from Stack-Exchange.\\
12 What different member statuses does H\&M have?\\
13 Tell me 3 Formula 1 drivers from Germany.\\
14 How many users does Avito have?\\
15 Give me the title of the oldest medical study\\
16 Tell me the 8 most expensive products from H\&M\\
17 Show me the AccountId of the longest existing user on Stack-Exchange.\\
18 How many customers at H\&M are younger than 40?\\
19 What is the cheapest advertising on Avito?\\
20 Which country has the most Formula 1 drivers?\\
\end{framed}

\section{Handling of Infinite Reasoning Loops}
\label{Handling of Infinite Reasoning Loops}

A key limitation of our agentic system is the potential for infinite reasoning loops. In such cases, the model repeatedly issues the same requests without acquiring any new information, effectively becoming stuck. For instance, it may continuously request the same metadata for a given table without progressing further.

Attempts to prevent the model from repeating requests or to encourage it to seek alternative information were not successful.

\textbf{Mitigation Strategy:}

\textbf{Forced Response after 10 Requests:} To ensure progress, the system is designed to force the model to provide an SQL query after 10 repeated requests. This intervention breaks the loop while maintaining the autonomy of the agent.

This approach provides a practical safeguard against infinite loops and ensures continued evaluation of the agentic system.

\end{document}